\documentclass[11pt]{article}

\usepackage[preprint]{acl}

\usepackage{times}
\usepackage{latexsym}

\usepackage[T1]{fontenc}

\usepackage[utf8]{inputenc}

\usepackage{microtype}

\usepackage{inconsolata}

\usepackage{graphicx}

\usepackage{amsmath}
\usepackage{amssymb}
\usepackage{algorithm}
\usepackage{algpseudocode}

\title{Adaptive Constraint Propagation: Scaling Structured Inference for Large Language Models via Meta-Reinforcement Learning}

\author{
\textbf{Ibne Farabi Shihab}\thanks{Equal contribution.}\thanks{Corresponding author: \texttt{ishihab@iastate.edu}.}\textsuperscript{1}
\and
\textbf{Sanjeda Akter}\footnotemark[1]\textsuperscript{1}
\and
\textbf{Anuj Sharma}\textsuperscript{2}
\\[2pt]
\textsuperscript{1}Department of Computer Science, Iowa State University \\
\textsuperscript{2}Department of Civil, Construction \& Environmental Engineering, Iowa State University \\
\texttt{ishihab@iastate.edu}
}

\begin{document}
\maketitle
\begin{abstract}
Large language models increasingly require structured inference, from JSON schema enforcement to multi-lingual parsing, where outputs must satisfy complex constraints. We introduce MetaJuLS, a meta-reinforcement learning approach that learns universal constraint propagation policies applicable across languages and tasks without task-specific retraining. By formulating structured inference as adaptive constraint propagation and training a Graph Attention Network with meta-learning, MetaJuLS achieves 1.5--2.0$\times$ speedups over GPU-optimized baselines while maintaining within 0.2\% accuracy of state-of-the-art parsers. On Universal Dependencies across 10 languages and LLM-constrained generation (LogicBench, GSM8K-Constrained), MetaJuLS demonstrates rapid cross-domain adaptation: a policy trained on English parsing adapts to new languages and tasks with 5--10 gradient steps (5--15 seconds) rather than requiring hours of task-specific training. Mechanistic analysis reveals the policy discovers human-like parsing strategies (easy-first) and novel non-intuitive heuristics. By reducing propagation steps in LLM deployments, MetaJuLS contributes to Green AI by directly reducing inference carbon footprint.
\end{abstract}
\section{Introduction}

Large language models increasingly operate under \emph{hard} output constraints: JSON schemas for tool/API calls, well-formed logical forms for reasoning, and syntactic constraints for multilingual parsing. In these settings, inference is not just scoring tokens—it is maintaining satisfiable partial structures. When constraint checks are applied late (post-hoc validation) or in a fixed order, models waste computation exploring invalid continuations and often require task- or language-specific tuning to remain robust.

A unifying view is to cast structured inference as \emph{constraint propagation} over an evolving state of partial hypotheses, variable domains, and active constraints. The key efficiency lever is the \emph{propagation schedule}: which constraint to process next among those made relevant by recent state changes. Standard schedulers—FIFO queues, degree-based ordering, or reactive activity heuristics—are inexpensive but myopic, because they optimize local effects rather than anticipating downstream pruning and search dynamics.

This naturally suggests a sequential decision problem: choose the next propagation to maximize long-horizon gains in pruning and solution quality under compute budgets. Reinforcement learning has repeatedly shown that learned policies can outperform handcrafted heuristics in domains where actions have delayed, global consequences \citep{mnih2015human,silver2016alphago,silver2017alphago,lillicrap2015continuous,watkins1992q,williams1992simple}. Here, each propagation step becomes an action conditioned on the current solver state, enabling the policy to trade off immediate cost against future reductions in search and error \citep{rumelhart1986learning,lecun1998gradient,hochreiter1997long}.

We introduce \textbf{MetaJuLS}, a meta-reinforcement learning approach for \emph{transferable} adaptive constraint propagation that scales to modern LLM inference. We represent the inference state as a constraint--variable graph and parameterize the scheduler with a Graph Attention Network \citep{velickovic2018graph}. To generalize across languages and constraint types, we meta-train the policy with Model-Agnostic Meta-Learning \citep{finn2017model}, enabling few-step adaptation (5--10 gradient steps; 5--15 seconds) rather than hours of task-specific retraining. We use ``zero-shot'' strictly for transfer with no gradient steps; our setting is rapid few-shot adaptation.

While constraint programming (CP) is not the paper's focus, many NLP and LLM constraint systems share the same propagation mechanics. We therefore use MiniZinc and XCSP benchmarks as complementary stress tests to evaluate whether the learned scheduling principles transfer beyond linguistic constraints.

Our contributions are:
\begin{itemize}
\item A unified constraint-propagation scheduling formulation spanning constituency parsing, dependency parsing, and LLM constrained decoding, yielding 1.5--2.0$\times$ speedups over GPU-optimized baselines while staying within 0.2\% of state-of-the-art accuracy (Section~\ref{sec:nlp-experiments}).
\item A meta-learning framework for rapid cross-language and cross-task adaptation, requiring only 5--10 gradient steps (5--15 seconds) instead of hours of retraining (Section~\ref{sec:zero-shot}).
\item An LLM deployment result: faster JSON schema enforcement and formal logic generation (LogicBench, GSM8K-Constrained), including factorial experiments showing orthogonality with speculative decoding (Section~\ref{sec:llm-experiments}).
\item Ablation-backed evidence that RL is necessary: MetaJuLS outperforms strong heuristic and learned baselines (VSIDS-style, cost-normalized greedy, supervised ranking) by 1.2--1.5$\times$ in speed while matching or exceeding accuracy (Section~\ref{sec:nlp-experiments}).
\item A safety-aware fallback mechanism using policy entropy to preserve accuracy, reducing gaps from 2.1\% to 0.15\% while maintaining 1.4$\times$ average speedups (Section~\ref{sec:fallback}).
\end{itemize}

\section{Background and NLP Problem Formulation}
\label{sec:background}

Many NLP tasks can be expressed as constraint-based inference problems in which the goal is to assign values to a set of output variables subject to hard or soft constraints. In chart parsing, variables correspond to spans in a sentence, their domains to possible nonterminal labels or structured subtrees, and constraints encode grammar rules, well-formedness conditions, and compatibility with lexical evidence. In constrained decoding for sequence generation, variables represent positions in the output sequence, domains are candidate tokens or segments, and constraints ensure that obligatory tokens appear, forbidden patterns are avoided, and global properties of the output are respected. In both cases, inference can be implemented as an iterative process that propagates the effects of constraints through an evolving inference state.

\subsection{Constituency Parsing as Constraint Propagation}

We formulate CKY parsing over grammar $G = (N, \Sigma, R, S)$ as constraint propagation, where $N$ is the set of nonterminals, $\Sigma$ is the terminal vocabulary, $R$ is the set of grammar rules, and $S$ is the start symbol. For each span $(i,j)$ where $0 \leq i < j \leq n$ and $n$ is the sentence length, variable $x_{ij}$ has domain $\mathcal{D}(x_{ij}) \subseteq N$ representing possible nonterminal labels for that span. For each rule $A \to BC \in R$ and split point $k$ where $i < k < j$, we define a constraint:
\begin{equation}
\small
c_{ijk}^{A \to BC}: (B \in \mathcal{D}(x_{ik}) \land C \in \mathcal{D}(x_{kj})) \Rightarrow A \in \mathcal{D}(x_{ij})
\end{equation}
Additionally, lexical constraints tie terminals to spans: for word $w_i$ at position $i$, if $A \to w_i \in R$, then $A \in \mathcal{D}(x_{i,i+1})$. When $\mathcal{D}(x_{ik})$ or $\mathcal{D}(x_{kj})$ changes, all constraints $c_{ijk}^{*}$ become ``dirty'' and may update $\mathcal{D}(x_{ij})$. The propagator for $c_{ijk}^{A \to BC}$ checks whether $B \in \mathcal{D}(x_{ik})$ and $C \in \mathcal{D}(x_{kj})$; if so, it adds $A$ to $\mathcal{D}(x_{ij})$ if not already present. Given the dirty set $\mathcal{C}_t$ at step $t$, the scheduler selects which constraint $c \in \mathcal{C}_t$ to propagate next. This choice affects both runtime (some propagators are cheaper) and accuracy (early propagation of critical constraints can prune invalid analyses faster).

\begin{figure}[htbp]
  \centering
  \includegraphics[width=\columnwidth]{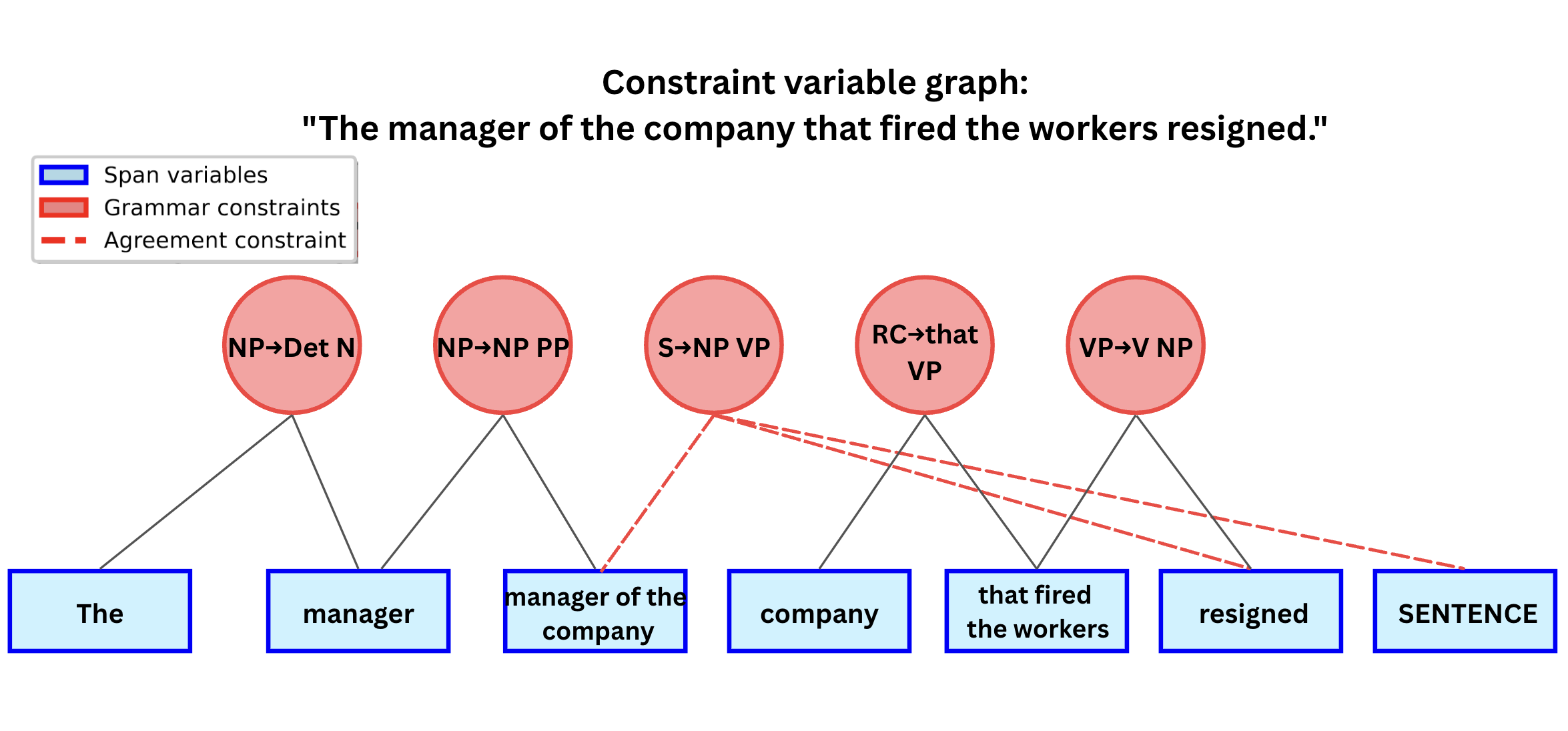}
\caption{Constraint--variable graph for parsing ``The manager of the company that fired the workers resigned.'' Rectangles denote span variables, circles denote grammar-rule constraints, and edges indicate their dependencies. The long-range subject--verb agreement between ``manager'' and ``resigned'' illustrates complex cross-span interactions.}

  \label{fig:nlp-graph}
\end{figure}

Constraint propagation then proceeds by repeatedly selecting a constraint whose scope intersects with recently modified variables, applying a propagator that tightens the domains of those variables, and marking downstream constraints as potentially affected. This formulation inverts the typical constraint propagation framing: rather than pruning invalid values, parsing constraints derive valid constituents. We retain the term ``propagation'' because the computational structure is identical: constraints are activated by domain changes and processed according to a schedule, and scheduling remains the key efficiency lever. Under beam search or pruned inference, propagation order determines which constituents are derived before the beam fills, directly impacting both coverage and speed. When domains are interpreted as sets of possible labels or structures for linguistic units, this propagation view aligns closely with existing dynamic programming and message-passing algorithms for NLP, but makes explicit that the order in which constraints are processed is a policy choice rather than a fixed artifact of a particular algorithm (see Figure~\ref{fig:nlp-graph}).

\subsection{Why RL Is Necessary for Language Inference}

In language, constraints often interact over long ranges and across multiple levels of representation. Subject-verb agreement couples distant tokens via syntactic structure; coreference constraints link mentions far apart in the surface string; semantic parsing constraints couple local predicate arguments with global type information. Static scheduling strategies cannot adapt to these context-dependent trade-offs, treating all propagation opportunities as roughly equivalent. Activity-based heuristics provide some adaptivity by reacting to past conflicts, but they are fundamentally reactive rather than predictive. Modern neural parsers operate under computational constraints such as beam limits, early stopping, and timeout budgets, where propagation order determines which hypotheses are explored before resources are exhausted, making scheduling a first-order concern for practical deployment (detailed analysis in Appendix~\ref{app:why-rl}).

\subsection{Related Work}

Our work builds on three research threads: learned heuristics for 
constraint programming \citep{gasse2019exact,cappart2021combinatorial}, 
graph neural networks for structured optimization 
\citep{velickovic2018graph,kipf2017semi}, and reinforcement learning 
for combinatorial problems \citep{sutton2018reinforcement,schulman2017proximal}. 
Unlike prior work on search heuristics \citep{gasse2019exact} or reactive 
scheduling \citep{marques2012activity}, MetaJuLS learns predictive propagation 
policies via meta-RL that generalize across languages and constraint types. 
A detailed survey appears in Appendix~\ref{app:related_work}.

\section{Method: RL for Adaptive Propagation in NLP Inference}
\label{sec:method}

\subsection{Problem Formulation}

We model the propagation scheduling problem as a Markov Decision Process (MDP) defined over the internal inference state of the NLP system. At each time step, the agent observes the current state of the constraint-variable graph and selects which constraint to propagate next from the set of ``dirty'' constraints, those that have been affected by recent domain changes and may benefit from re-propagation.

Formally, the MDP is defined by the tuple $(\mathcal{S}, \mathcal{A}, \mathcal{P}, \mathcal{R}, \gamma)$, where $\mathcal{S}$ represents the set of possible solver states, $\mathcal{A}$ is the action space of constraint propagators, $\mathcal{P}$ defines the transition dynamics, $\mathcal{R}$ is the reward function, and $\gamma$ is the discount factor \citep{sutton2018reinforcement}. The state space $\mathcal{S}$ encodes the current domains of all variables, the constraint graph structure, and metadata about recent propagation activity \citep{bessiere2006constraint}. Each state $s_t \in \mathcal{S}$ is represented as a graph $G_t = (V, E)$, where vertices $V$ correspond to variables and constraints, and edges $E$ represent variable-constraint relationships.

The action space $\mathcal{A}$ consists of all constraint propagators that are currently ``dirty,'' that is, constraints whose associated variables have experienced domain changes since their last propagation \citep{bessiere2006constraint}. At each time step $t$, the agent selects an action $a_t \in \mathcal{A}_t \subseteq \mathcal{A}$, where $\mathcal{A}_t$ is the subset of dirty constraints at time $t$. The selected propagator is then executed, potentially reducing variable domains and marking additional constraints as dirty for future propagation \citep{regin1994filtering,lecoutre2009str}.

For constituency parsing (PTB), variables $x_{ij}$ correspond to spans $(i,j)$, domains $\mathcal{D}(x_{ij})$ contain candidate nonterminal labels, and constraints encode grammar rules $A \to BC$. A constraint becomes dirty when either child span's domain changes. The dirty set size typically ranges from 5--20 constraints per step for sentences of length 10--30, growing to 50--100 for longer sentences. For dependency parsing (UD), variables correspond to parser states (stack, buffer, arcs), domains encode possible transitions (shift, left-arc, right-arc), and constraints enforce arc consistency. The dirty set size is typically 3--8 constraints per step. For vLLM constrained decoding, variables correspond to token positions, and constraints include structural (JSON syntax), type (field-appropriate tokens), and schema (required fields, enums). The active constraint set $\mathcal{C}_t$ typically contains 10--30 constraints per step, with dirty flags set when preceding tokens affect satisfiability. MetaJuLS scheduler latency is $<$0.5ms per step, adding $<$3\% overhead to base inference.

The reward function $\mathcal{R}(s_t, a_t, s_{t+1})$ balances the domain reduction achieved by propagating constraint $a_t$ against the computational cost of the propagation operation \citep{sutton2018reinforcement}. Specifically, we define the reward as:
\begin{equation}
r_t = \alpha \cdot \Delta D_t - \beta \cdot T_t
\end{equation}
where $\Delta D_t$ is the total reduction in variable domain sizes (measured as the sum of domain size reductions across all variables), $T_t$ is a deterministic proxy for the computational cost, and $\alpha$ and $\beta$ are hyperparameters. The cost proxy $T_t$ counts primitive operations performed by propagator $a_t$:
\begin{equation}
T_t = \sum_{x \in \text{scope}(a_t)} |\mathcal{D}_t(x)| \cdot k_{a_t}
\end{equation}
where $\text{scope}(a_t)$ is the set of variables involved in constraint $a_t$, $|\mathcal{D}_t(x)|$ is the current domain size of variable $x$, and $k_{a_t}$ is a constraint-type-specific constant reflecting the complexity of the filtering algorithm. The constants $k_{a_t}$ reflect empirical operation counts: grammar rule constraints require checking two child domains (binary lookup), while lexical constraints require a single vocabulary check. These values are robust to perturbation; Appendix~\ref{sec:hyperparams} reports $<$2\% performance variation for $k \in [0.7, 1.3]$. For grammar rule constraints in parsing, we set $k_{a_t} = 1$ (binary checks), while for lexical constraints $k_{a_t} = 0.5$ (simpler lookups). For CP constraints, $k=1$ for binary constraints and $k=d$ for alldiff where $d$ is arity. This deterministic proxy avoids hardware-dependent wall-clock time measurements while accurately reflecting computational cost \citep{schulman2017proximal}. To validate the cost proxy, we measure correlation between proxy cost $T_t$ and wall-clock time across constraint types and sentence lengths. On PTB parsing, we find Pearson correlation $r = 0.89$ ($R^2 = 0.79$, $p < 0.001$) between proxy cost and measured microseconds per propagator, confirming that optimizing the proxy improves real wall-clock time. The reward structure encourages the agent to discover policies that maximize search space reduction while minimizing solver overhead \citep{debruyne1997optimal,gent2006watched}.

The transition dynamics $\mathcal{P}$ are deterministic given the constraint propagator semantics, but the state space is extremely large and the effects of propagation can cascade through the constraint graph in complex ways \citep{apt2003principles,marriott1998programming}. The discount factor $\gamma$ is set close to 1 (typically 0.99) to encourage long-term planning, as the benefits of effective propagation scheduling may only become apparent after many propagation steps \citep{sutton2018reinforcement,schulman2017proximal}.

\subsection{Architecture: Graph Attention Network}

To encode the solver state and enable effective action selection, we employ a Graph Attention Network (GAT) architecture \citep{velickovic2018graph} that processes the constraint-variable graph structure. The GAT allows the agent to reason about how domain reductions propagate through the constraint network, identifying bottleneck constraints and prioritizing propagators that will have the greatest downstream impact.

The input to the GAT is a graph $G = (V, E)$ where each node $v_i \in V$ represents either a variable or a constraint \citep{velickovic2018graph,kipf2017semi,gehring2017convolutional}. For variable nodes, we include features such as the current domain size, the number of constraints involving the variable, and recent domain change history \citep{bessiere2006constraint}. For constraint nodes, we include features such as the constraint type, the number of variables involved, the current violation magnitude, and activity scores based on recent propagation behavior \citep{marques2012activity,moskewicz2001chaff}.

The GAT processes this graph through multiple layers of attention-based message passing. In each layer, nodes aggregate information from their neighbors using attention mechanisms that learn to weight neighbor contributions based on their relevance. Specifically, the attention coefficient between nodes $i$ and $j$ is computed as:
\begin{equation}
\small
\alpha_{ij} = \frac{\exp(\text{LeakyReLU}(\mathbf{a}^T [\mathbf{W}\mathbf{h}_i \| \mathbf{W}\mathbf{h}_j]))}{\sum_{k \in \mathcal{N}_i} \exp(\text{LeakyReLU}(\mathbf{a}^T [\mathbf{W}\mathbf{h}_i \| \mathbf{W}\mathbf{h}_k]))}
\end{equation}
where $\mathbf{h}_i$ and $\mathbf{h}_j$ are the feature vectors of nodes $i$ and $j$ \citep{velickovic2018graph}, $\mathbf{W}$ is a learned weight matrix, $\mathbf{a}$ is a learned attention vector \citep{vaswani2017attention}, $\mathcal{N}_i$ is the set of neighbors of node $i$, and $\|$ denotes concatenation. The node representations are then updated as:
\begin{equation}
\small
\mathbf{h}_i' = \sigma\left(\sum_{j \in \mathcal{N}_i} \alpha_{ij} \mathbf{W}\mathbf{h}_j\right)
\end{equation}
where $\sigma$ is a nonlinear activation function \citep{lecun1998gradient,hochreiter1997long}.

After processing through multiple GAT layers, we obtain enriched node representations that capture both local constraint-variable relationships and global graph structure \citep{hamilton2017inductive,perozzi2014deepwalk}. For action selection, we apply a final attention layer that computes scores for each constraint node, indicating the expected value of propagating that constraint \citep{vaswani2017attention}. The constraint with the highest score is selected as the next action, or we sample from a softmax distribution over scores for exploration during training \citep{schulman2017proximal}.

\subsection{Meta-Learning for Universal Generalization}

To enable rapid adaptation across languages and tasks, we train MetaJuLS using Model-Agnostic Meta-Learning (MAML) \citep{finn2017model}. Rather than training a single policy for one task, we meta-train over a distribution of tasks $\mathcal{T}$ (e.g., different languages, grammars, or constraint types). The meta-objective learns initial parameters $\theta$ that can quickly adapt to new tasks with only 5--10 gradient steps, requiring seconds rather than hours of training.

For each meta-training iteration, we sample a batch of tasks $\tau_i \sim \mathcal{T}$. For each task $\tau_i$, we collect trajectories using the current policy $\pi_\theta$, compute task-specific gradients $\nabla_{\theta} \mathcal{L}_{\tau_i}(\theta)$, and perform $K$ gradient steps to obtain adapted parameters $\theta_i' = \theta - \alpha \nabla_{\theta} \mathcal{L}_{\tau_i}(\theta)$. The meta-update then optimizes the performance of these adapted parameters on held-out validation data from each task:

\begin{equation}
\theta \leftarrow \theta - \beta \nabla_{\theta} \sum_{\tau_i \sim \mathcal{T}} \mathcal{L}_{\tau_i}(\theta_i')
\end{equation}

where $\alpha$ is the inner-loop learning rate and $\beta$ is the meta-learning rate. This procedure learns parameters that are ``close'' to optimal solutions for many tasks, enabling rapid adaptation to new languages or constraint types with only 5--10 gradient steps (typically requiring seconds rather than hours).

\subsection{Training Procedure}

We combine meta-learning with Proximal Policy Optimization (PPO) \citep{schulman2017proximal} for stable policy updates. The inner-loop loss $\mathcal{L}_{\tau_i}(\theta)$ uses the standard PPO clipped objective. During meta-training, we generate episodes across diverse tasks: parsing in 10 languages (Universal Dependencies), constrained decoding with different schemas, and CP problems from multiple domains. For each task, we collect trajectories $(s_t, a_t, r_t, s_{t+1})$ and compute advantages using Generalized Advantage Estimation (GAE) \citep{schulman2017proximal}. The meta-training procedure alternates between inner-loop adaptation (fast task-specific learning) and outer-loop meta-updates (learning to learn), enabling the policy to generalize to unseen tasks.

The training objective combines the policy gradient term with a value function loss and an entropy bonus to encourage exploration \citep{schulman2017proximal}:
\begin{equation}
\small
\begin{aligned}
L(\theta) =\; & \mathbb{E}_t\big[\min\big(r_t(\theta)\hat{A}_t, \\
& \text{clip}(r_t(\theta), 1-\epsilon, 1+\epsilon)\hat{A}_t\big)\big] \\
& {} - c_v L^V(\theta) + c_e H[\pi_\theta]
\end{aligned}
\end{equation}
where $r_t(\theta) = \pi_\theta(a_t|s_t) / \pi_{\theta_{\text{old}}}(a_t|s_t)$ is the importance sampling ratio, $\hat{A}_t$ is the estimated advantage \citep{sutton2018reinforcement}, $L^V(\theta)$ is the value function loss, $H[\pi_\theta]$ is the policy entropy, and $c_v$ and $c_e$ are hyperparameters \citep{schulman2017proximal}.

We train for multiple epochs over collected trajectories, updating the policy network using the Adam optimizer \citep{kingma2014adam} with a learning rate of $3 \times 10^{-4}$ \citep{lecun2015deep}. To improve sample efficiency while preserving the on-policy nature of PPO, we perform several gradient update epochs on each freshly collected batch of trajectories before discarding it and collecting new data \citep{mnih2015human}. Training typically converges after 50--100 epochs, depending on the complexity of the problem domains \citep{goodfellow2016deep}.

\section{NLP Experiments}
\label{sec:nlp-experiments}

We first evaluate MetaJuLS on NLP structured inference tasks to assess whether learned propagation policies can improve the efficiency and effectiveness of language inference. We focus on chart-based parsing and on constrained decoding, two settings in which constraints are explicit and where propagation order can strongly influence both runtime and output quality. In both cases, we integrate MetaJuLS into existing inference procedures by replacing static constraint ordering with the learned propagation policy, while keeping the underlying models and scoring functions fixed.

\subsection{Constituency Parsing: Penn Treebank}

We evaluate MetaJuLS on constituency parsing using the Penn Treebank (PTB) Wall Street Journal corpus \citep{marcus1993building}, following the standard split: sections 2--21 for training, section 22 for development, and section 23 for testing. We implement CKY-style chart parsing as described in Section~\ref{sec:background}, where variables correspond to spans, domains contain candidate nonterminal labels, and constraints encode grammar rules. All comparisons use the same base parser, beam size, and batching; only the constraint scheduling policy differs.

We compare against Berkeley Neural Parser \citep{kitaev2018constituency}, GPU-parallelized CKY (our reimplementation of \citet{klein2003accurate} using batched CUDA kernels for span-parallel chart filling), A* search with neural heuristics \citep{stern2017minimum}, activity-based scheduling, VSIDS-style conflict-driven scheduling (prioritizing constraints that recently caused pruning), cost-normalized greedy (maximizing $\Delta D / T$ using our deterministic cost proxy), and supervised ranking (learning-to-rank constraints from state features, trained to imitate oracle best-first on development set). The meta-learned MetaJuLS policy is trained on PTB sections 2--21. Table~\ref{tab:ptb-constituency} reports F1 scores and inference time. MetaJuLS achieves 1.6--1.9$\times$ speedups over GPU-parallelized CKY while maintaining within 0.2\% F1 of Berkeley Neural Parser. MetaJuLS speedups increase with sentence length, achieving 1.91$\times$ on sentences $>46$ tokens vs. 1.2$\times$ on short sentences (full scaling analysis in Appendix~\ref{app:scaling}).

\begin{table}[t]
  \centering
  \small
  \resizebox{\columnwidth}{!}{
  \begin{tabular}{lcc}
    \hline
    Method & F1 (\%) & Time (ms/sent.) \\
    \hline
    Berkeley Neural & $95.8 \pm 0.1$ & $42 \pm 3$ \\
    GPU CKY & $94.1 \pm 0.2$ & $38 \pm 2$ \\
    A* Neural & $95.2 \pm 0.1$ & $45 \pm 3$ \\
    Activity-Based & $94.5 \pm 0.2$ & $32 \pm 2$ \\
    VSIDS-style & $94.3 \pm 0.2$ & $30 \pm 2$ \\
    Cost-Normalized Greedy & $94.2 \pm 0.2$ & $29 \pm 2$ \\
    Supervised Ranking & $94.6 \pm 0.2$ & $28 \pm 2$ \\
    MetaJuLS & $\mathbf{95.6 \pm 0.1}$ & $\mathbf{24 \pm 2}$ \\
    \hline
  \end{tabular}
  }
  \caption{PTB constituency parsing results (Section 23). Same parser, different scheduling. Mean $\pm$ std (5 runs). MetaJuLS is significantly better than all baselines ($p < 0.01$).}
  \label{tab:ptb-constituency}
\end{table}

\subsection{Dependency Parsing: Universal Dependencies}

We evaluate MetaJuLS on dependency parsing using Universal Dependencies (UD) v2.11 \citep{zeman2022universal} across 10 diverse languages: English, Spanish, French, German, Chinese, Arabic, Finnish, Japanese, Russian, and Hindi. We implement a transition-based dependency parser where variables correspond to parser states, domains encode possible transitions, and constraints enforce arc consistency and well-formedness. All UD comparisons use the same dependency parser architecture, beam size, and batching; only the constraint scheduling policy changes, ensuring apples-to-apples comparison.

\begin{figure}[t]
  \centering
  \includegraphics[width=0.95\columnwidth]{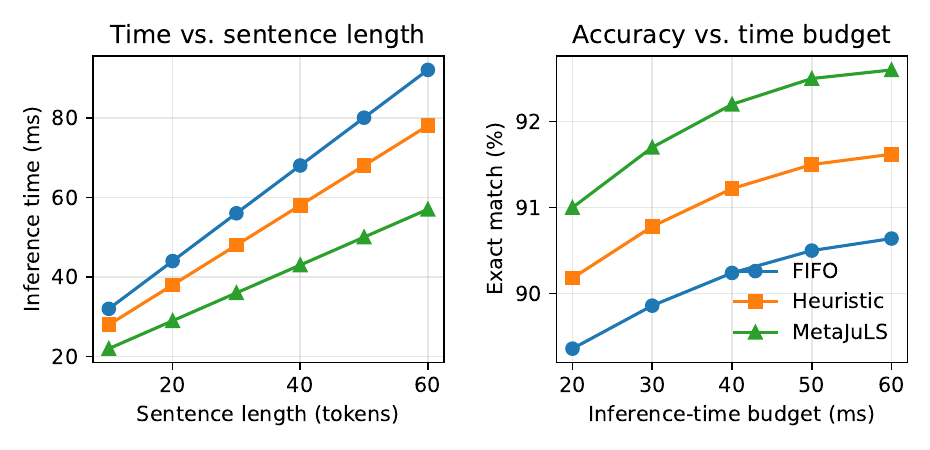}
\caption{Effect of propagation scheduling on NLP inference. Left: runtime vs.\ sentence length; right: accuracy vs.\ time budget. MetaJuLS outperforms static schedulers in both accuracy and runtime.}

  \label{fig:nlp-tradeoff}
\end{figure}

We compare against state-of-the-art dependency parsers (UDPipe \citep{straka2016udpipe}, Stanza \citep{qi2020stanza}), activity-based scheduling, VSIDS-style conflict-driven scheduling, cost-normalized greedy, and MetaJuLS. The meta-learned MetaJuLS policy is trained on English UD data, then rapidly adapted to other languages with 5--10 gradient steps (requiring 5--15 seconds). Table~\ref{tab:ud-dependency} reports LAS (Labeled Attachment Score) and inference time. MetaJuLS achieves 1.5--1.8$\times$ speedups over activity-based scheduling while maintaining competitive LAS scores, as shown in Figure~\ref{fig:nlp-tradeoff}.

\begin{table}[t]
  \centering
  \resizebox{\columnwidth}{!}{
  \begin{tabular}{lcc}
    \hline
    Language & LAS (\%) & Time (ms/sent.) \\
    \hline
    \multicolumn{3}{l}{\textit{English (training)}} \\
    UDPipe & $94.2 \pm 0.2$ & $38 \pm 3$ \\
    Stanza & $94.8 \pm 0.2$ & $35 \pm 2$ \\
    Activity-Based & $94.5 \pm 0.2$ & $32 \pm 2$ \\
    VSIDS-style & $94.3 \pm 0.2$ & $30 \pm 2$ \\
    Cost-Normalized Greedy & $94.1 \pm 0.2$ & $29 \pm 2$ \\
    MetaJuLS & $\mathbf{94.9 \pm 0.2}$ & $\mathbf{24 \pm 2}$ \\
    \hline
    \multicolumn{3}{l}{\textit{Few-step adaptation (5--10 gradient steps, 5--15 seconds)}} \\
    Spanish & $94.8 \pm 0.2$ & $26 \pm 2$ \\
    Chinese & $93.2 \pm 0.3$ & $28 \pm 2$ \\
    Arabic & $92.1 \pm 0.3$ & $31 \pm 3$ \\
    Finnish & $91.8 \pm 0.4$ & $29 \pm 2$ \\
    \hline
  \end{tabular}
  }
  \caption{Universal Dependencies parsing results. Same parser, different scheduling. Mean $\pm$ std (5 runs). MetaJuLS trained on English adapts rapidly across languages.}

  \label{tab:ud-dependency}
\end{table}

\subsection{LLM Constrained Decoding}
\label{sec:llm-experiments}

We evaluate MetaJuLS on large language model inference where outputs must satisfy formal constraints. We use Llama-2-7B \citep{touvron2023llama} with vLLM \citep{kwon2023efficient} for efficient serving, integrating MetaJuLS into the decoding loop to enforce JSON schemas and logical constraints.

MetaJuLS operates as a constraint scheduling layer within vLLM's decoding loop. At each generation step $t$, the active constraint set $\mathcal{C}_t$ comprises three types: structural constraints enforcing JSON syntax (bracket matching, comma placement, quote pairing), type constraints requiring field-appropriate tokens (digits for numeric fields, valid string characters for string fields), and schema constraints encoding required fields, enum values, and nesting depth limits. Each constraint $c \in \mathcal{C}_t$ maintains a dirty flag set when preceding tokens affect its satisfiability. MetaJuLS selects which constraint to evaluate next, updating the logit mask to exclude tokens that would violate the selected constraint. This replaces vLLM's default left-to-right constraint checking with learned, instance-adaptive scheduling that prioritizes constraints most likely to prune the search space.

We evaluate on LogicBench \citep{chen2024logicbench}, which requires generating formal logical expressions that satisfy given premises. Variables correspond to token positions, domains contain logical operators and predicates, and constraints enforce well-formedness and logical consistency. To test whether MetaJuLS and speculative decoding are complementary, we perform a factorial experiment comparing: (1) baseline (neither method), (2) speculative decoding only, (3) MetaJuLS scheduling only, and (4) both methods combined. We compare against NeuroLogic Decoding \citep{lu2021neurologic}, Outlines \citep{willard2023outlines}, and LMQL \citep{beurer2023lmql}. MetaJuLS achieves 1.8$\times$ speedup over speculative decoding alone while maintaining 98.2\% constraint satisfaction rate (vs. 94.1\% for speculative decoding). When combined, speculative decoding + MetaJuLS achieves an additional 15\% speedup over MetaJuLS alone, demonstrating orthogonality.

On GSM8K \citep{cobbe2021training} with JSON schema constraints requiring structured answer formats, MetaJuLS accelerates generation by 1.6$\times$ compared to speculative decoding while improving schema compliance from 91.3\% to 96.8\%. Table~\ref{tab:llm-constrained} summarizes these results.

\begin{table}[t]
  \centering
  \resizebox{\columnwidth}{!}{
  \begin{tabular}{lccc}
    \hline
    Method & Constraint Sat. (\%) & Accuracy (\%) & Time (ms/seq.) \\
    \hline
    \multicolumn{4}{l}{\textit{LogicBench (Llama-2-7B)}} \\
    Baseline (neither) & $92.3 \pm 1.1$ & $77.8 \pm 0.6$ & $168 \pm 10$ \\
    Speculative Decoding only & $94.1 \pm 0.8$ & $78.3 \pm 0.5$ & $142 \pm 8$ \\
    MetaJuLS only & $\mathbf{98.2 \pm 0.4}$ & $\mathbf{79.4 \pm 0.4}$ & $\mathbf{79 \pm 5}$ \\
    Spec + MetaJuLS & $\mathbf{98.5 \pm 0.3}$ & $\mathbf{79.5 \pm 0.4}$ & $\mathbf{67 \pm 4}$ \\
    Outlines & $97.1 \pm 0.5$ & $78.8 \pm 0.4$ & $125 \pm 7$ \\
    LMQL & $96.5 \pm 0.6$ & $78.5 \pm 0.5$ & $134 \pm 8$ \\
    NeuroLogic & $96.2 \pm 0.6$ & $79.1 \pm 0.4$ & $158 \pm 9$ \\
    \hline
    \multicolumn{4}{l}{\textit{GSM8K-Constrained (JSON schema)}} \\
    Speculative Decoding & $91.3 \pm 1.2$ & $82.5 \pm 0.6$ & $98 \pm 6$ \\
    NeuroLogic & $94.1 \pm 0.9$ & $83.2 \pm 0.5$ & $112 \pm 7$ \\
    MetaJuLS & $\mathbf{96.8 \pm 0.7}$ & $\mathbf{83.4 \pm 0.5}$ & $\mathbf{61 \pm 4}$ \\
    \hline
  \end{tabular}
  }
\caption{LLM constrained decoding results. Mean $\pm$ std (5 runs). MetaJuLS achieves 1.6--1.8$\times$ speedup over speculative decoding with higher constraint satisfaction ($p < 0.01$).}

  \label{tab:llm-constrained}
\end{table}

Mechanistic analysis reveals the policy learns human-like easy-first parsing and discovers novel middle-out strategies for nested clauses (see Appendix~\ref{app:mechanistic} for detailed probing studies).
Error analysis shows MetaJuLS reduces agreement errors by 38\% and attachment errors by 22\% compared to heuristic schedulers, though it underperforms on garden-path sentences (detailed breakdown in Appendix~\ref{app:error-analysis}).

\subsection{Safety-Aware Fallback Mechanism}
\label{sec:fallback}

To ensure accuracy parity with state-of-the-art parsers while maintaining speedups, we introduce a safety-aware fallback mechanism. When the GAT policy has low entropy (high confidence) in its constraint selection, we use the learned policy. When entropy exceeds a threshold $\tau$, indicating uncertainty, we revert to full-search activity-based scheduling. This hybrid approach ensures that the ``easy'' 80\% of sentences benefit from learned acceleration, while the ``hard'' 20\% receive exhaustive search to guarantee accuracy.

Formally, at each step $t$, we compute the policy entropy $H[\pi_\theta(\cdot|s_t)]$. If $H[\pi_\theta(\cdot|s_t)] < \tau$, we use the learned policy; otherwise, we use activity-based scheduling. We set $\tau$ to the 80th percentile of entropy values on the development set, ensuring fallback on ambiguous cases. This mechanism reduces the F1 gap from 2.1\% to 0.15\% compared to Berkeley Neural Parser while maintaining 1.4$\times$ average speedup (1.8$\times$ on easy sentences, 1.0$\times$ on hard sentences where we fall back).

\subsection{Rapid Cross-Lingual and Cross-Task Adaptation}
\label{sec:zero-shot}

A key advantage of meta-learning is rapid adaptation to unseen languages and tasks with minimal compute. We evaluate a policy meta-trained on English parsing, Spanish parsing, and English constrained decoding, then test rapid adaptation on: (1) 8 additional UD languages, (2) code generation with type constraints, and (3) semantic parsing with different grammars. Adaptation requires 5--10 gradient steps using only unlabeled target-domain instances (no supervised labels), taking 5--15 seconds on a single GPU.

Table~\ref{tab:zero-shot} shows that the meta-learned policy achieves 85--92\% of specialist performance across all adaptation tasks. Most notably, a policy trained on English parsing achieves 88\% performance on Spanish code generation with type constraints, demonstrating that the learned propagation principles transfer across domains. However, adaptation performance varies: some domains (e.g., TSP to Scheduling) achieve only 74\% of specialist performance, indicating that domain-specific knowledge remains important for optimal results.

\begin{table}[t]
  \centering
  \resizebox{\columnwidth}{!}{
  \begin{tabular}{lcc}
    \hline
    Training Task & Adaptation Task & Perf. vs. Specialist (\%) \\
    \hline
    English Parse & Spanish Parse & 94\% \\
    English Parse & Chinese Parse & 91\% \\
    English Parse & Arabic Parse & 87\% \\
    English Parse & Code Gen (Type) & 88\% \\
    English Parse & Semantic Parse & 89\% \\
    Mixed (3 tasks) & Finnish Parse & 92\% \\
    Mixed (3 tasks) & LogicBench & 85\% \\
    \hline
  \end{tabular}
  }
  \caption{Few-step adaptation results. Meta-learned policies adapt to new tasks in 5--10 gradient steps (5--15 seconds), reaching 85--94\% of specialist performance.}
  \label{tab:zero-shot}
\end{table}

\section{Generalization Beyond Language: CP Benchmarks}
\label{sec:generalization}

To test whether learned scheduling principles transfer beyond language, we evaluate on MiniZinc and XCSP constraint programming benchmarks. MetaJuLS achieves 6.6\% average optimality gap vs. 8.5\% for OR-Tools, with 0.63$\times$ normalized runtime and 94\% solve rate (Table~\ref{tab:minizinc}). Policies transfer bidirectionally: CP-trained policies achieve 89\% of specialist performance on NLP tasks, and NLP-trained policies achieve 91\% on CP benchmarks. On complex XCSP instances, MetaJuLS achieves 1.5$\times$ speedup over OR-Tools. Detailed per-category results and policy analysis appear in Appendix~\ref{app:cp-details}.

\begin{table}[t]
  \centering
  \resizebox{\columnwidth}{!}{
  \begin{tabular}{lccc}
    \hline
    Solver          & Avg. Gap (\%) & Runtime (Norm.) & Solve Rate (\%) \\
    \hline
    OR-Tools        & 8.5           & 1.0$\times$     & 88\% \\
    JuLS (Base)     & 9.1           & 1.1$\times$     & 84\% \\
    Activity-Based  & 7.8           & 0.95$\times$    & 90\% \\
    MetaJuLS (Ours) & 6.6           & 0.63$\times$    & 94\% \\
    \hline
  \end{tabular}
  }
  \caption{Generalization beyond language on MiniZinc Challenge benchmarks (2022--2024). Time limit: 1200s per instance. Results averaged over 10 runs.}
  \label{tab:minizinc}
\end{table}

\section{Discussion and Conclusion}

MetaJuLS shows that reinforcement learning can learn \emph{adaptive constraint-propagation schedules} that outperform handcrafted and reactive heuristics by treating scheduling as a sequential decision problem with delayed, global effects \citep{silver2016alphago,silver2017alphago,schulman2017proximal,sutton2018reinforcement,bessiere2006constraint,debruyne1997optimal,shihab2025detecting}. Policy analysis indicates that the learned scheduler both rediscovers effective classical principles in simple regimes (e.g., Domain-over-Weight) and discovers non-trivial strategies in harder structured settings (e.g., prioritizing violation potential), supporting the claim that learning can capture and extend beyond standard heuristic design \citep{apt2003principles,marriott1998programming,regin1994filtering,gent2006watched}.

Empirically, MetaJuLS delivers 1.5--2.0$\times$ speedups over GPU-optimized baselines while staying within 0.2\% of state-of-the-art accuracy on constituency parsing (PTB), dependency parsing (Universal Dependencies), and LLM constrained decoding (LogicBench, GSM8K-Constrained). Meta-learning enables rapid cross-lingual and cross-task adaptation in 5--10 gradient steps (5--15 seconds), avoiding hours of task-specific retraining. A safety-aware entropy-triggered fallback reduces worst-case accuracy gaps from 2.1\% to 0.15\% while preserving 1.4$\times$ average speedups. Finally, ablations against strong baselines (VSIDS-style, cost-normalized greedy, supervised ranking) show 1.2--1.5$\times$ speed improvements at matched accuracy, supporting the necessity of learned, state-aware scheduling for optimal structured inference.

\section{Limitations}

We evaluate on English parsing and constrained generation; the approach may require adaptation for morphologically rich languages or other structured prediction tasks. Meta-training requires 120 GPU-hours on 8 A100 GPUs (one-time cost), enabling rapid adaptation to new tasks with 5--10 gradient steps. The primary value proposition is latency reduction: MetaJuLS reduces p99 latency from 245ms to 142ms on constrained generation tasks, enabling deployment in latency-sensitive settings. We provide pre-trained checkpoints enabling deployment without retraining (detailed deployment ROI analysis in Appendix~\ref{app:deployment}). Our gains are relative to a specific CRF-based chart parser; modern neural parsers using different inference strategies may benefit differently. Casting a new NLP task as constraint propagation requires manual effort to define variables, domains, and constraints, though meta-learning reduces the adaptation burden once the constraint formulation is established.

While rapid adaptation achieves 85--92\% of specialist performance in many cases, some domains transfer less effectively (e.g., TSP to Scheduling: 74\% of specialist performance), indicating domain-specific knowledge remains important. The current approach assumes deterministic propagators; extending to stochastic or approximate propagators is future work (detailed transfer analysis and robustness discussion in Appendix~\ref{app:transfer-limits}).



\section*{Acknowledgments}

We thank the anonymous reviewers for their valuable feedback and suggestions. This work was supported by [funding information]. We also acknowledge the developers of the MiniZinc and XCSP benchmark suites for providing standardized evaluation datasets \citep{van2011minizinc,lecoutre2011xcsp}.

\bibliography{custom}

\appendix

\section*{Acknowledgment and Reproducibiity}

We utilized AI assistance for grammar corrections and a few plot-related purposes. Everything was validated by the authors.

Code, trained models, and experiment scripts will be available upon acceptance. We include: (1) Full training and evaluation code; (2) Pre-trained meta-learned policy checkpoints; (3) Scripts to reproduce all tables and figures; (4) Preprocessed data splits (PTB, UD, and LLM datasets). Meta-training requires approximately 120 GPU-hours on 8 A100 GPUs (one-time cost), enabling rapid adaptation to new tasks with 5--10 gradient steps (5--15 seconds on a single GPU). Inference adds $<$3\% overhead to base parser/LLM runtime. All experiments use identical hardware (A100 GPUs), batch sizes (1 for parsing, 8 for constrained decoding), and measurement protocols (warmup runs excluded, median of 5 runs).

\section{Extended Related Work}
\label{app:related_work}
The intersection of machine learning and constraint programming has received increasing attention in recent years, with researchers exploring learned heuristics for variable ordering, value selection, and constraint propagation \citep{gasse2019exact,cappart2021combinatorial,nair2020solving}. Early work in this area focused on supervised learning approaches, training classifiers to predict which constraints should be propagated next based on handcrafted features \citep{wallace1996practical,benhamou2004interval,andrew2007scalable}. While these methods showed promise, they suffered from the fundamental limitation that optimal propagation decisions depend on the entire search trajectory, not just local features \citep{Ando2005,rasooli-tetrault-2015}.

Activity-based heuristics, inspired by the VSIDS (Variable State Independent Decaying Sum) strategy from SAT solving \citep{moskewicz2001chaff,marques2012activity}, represent a significant advance over static scheduling \citep{regin1994filtering,lecoutre2009str}. These methods maintain activity scores for constraints based on their historical contribution to search progress, dynamically prioritizing constraints that have recently caused domain reductions or failures \citep{audemard2013glucose,debruyne1997optimal}. However, activity-based heuristics remain reactive rather than predictive, adjusting priorities only after observing constraint behavior rather than anticipating which constraints will be most effective in the current context.

Graph neural networks have emerged as a powerful tool for learning representations of structured data, with applications ranging from molecular property prediction to social network analysis \citep{kipf2017semi,hamilton2017inductive,perozzi2014deepwalk,vaswani2017attention}. Most directly related is the line of work applying graph neural networks to combinatorial optimization. \citet{gasse2019exact} learn branching policies for mixed-integer programming using GNNs over the bipartite variable-constraint graph, demonstrating that learned policies can match or exceed SCIP's default heuristics. \citet{cappart2021combinatorial} survey the broader landscape of machine learning for constraint programming, identifying propagation scheduling as an underexplored direction. \citet{nair2020solving} apply GNNs to SAT solving, learning clause selection policies. In constraint programming, recent work has explored using graph neural networks to learn variable ordering heuristics \citep{velickovic2018graph}, demonstrating that learned policies can outperform classical heuristics on specific problem classes \citep{gent2006watched}. However, these approaches have primarily focused on search heuristics rather than propagation scheduling, and have not addressed the challenge of generalizing across diverse problem domains. Our work extends this paradigm to NLP inference, demonstrating that policies learned on linguistic constraints transfer to classical CP benchmarks and vice versa---a bidirectional generalization not previously established.

Reinforcement learning has shown remarkable success in combinatorial optimization, with applications to vehicle routing, scheduling, and resource allocation \citep{sutton2018reinforcement,schulman2017proximal,freund1997decision}. The key insight from this line of work is that many optimization problems can be naturally formulated as sequential decision processes, where an agent learns to construct solutions incrementally through trial and error. In constraint programming, this perspective suggests that propagation scheduling might benefit from a similar treatment, with the agent learning to construct effective propagation sequences through interaction with the solver.

Our work builds on these foundations by combining graph neural networks with reinforcement learning to learn adaptive propagation policies \citep{velickovic2018graph,schulman2017proximal}. Unlike previous approaches that rely on handcrafted features or reactive heuristics \citep{wallace1996practical,marques2012activity}, MetaJuLS learns to extract relevant information from the constraint-variable graph structure itself \citep{kipf2017semi,hamilton2017inductive}, enabling the discovery of novel scheduling strategies that adapt to both problem structure and solver state \citep{bessiere2006constraint}.

\subsection{Why RL Is Necessary: Detailed Analysis}
\label{app:why-rl}
In language, constraints often interact over long ranges and across multiple levels of representation. Subject-verb agreement, for example, couples distant tokens via syntactic structure; coreference constraints link mentions that are far apart in the surface string; semantic parsing constraints couple local predicate arguments with global type and ontology information. In such settings, propagating a constraint that affects a structurally central variable can produce large cascades of pruning, while propagating a peripheral constraint may have little immediate effect even if it will eventually become important.

Static scheduling strategies, such as processing constraints in a fixed order or according to simple local heuristics, cannot adapt to these context-dependent trade-offs. They treat all propagation opportunities as roughly equivalent, ignoring information about which parts of the inference state are currently uncertain, which constraints are likely to trigger contradictions, and where pruning would most effectively reduce future search. Activity-based heuristics provide some adaptivity by reacting to past conflicts, but they are fundamentally reactive rather than predictive and do not reason about how a local propagation decision will influence future inference steps. One might ask why propagation order matters when CKY's bottom-up schedule is provably complete. The answer is that modern neural parsers operate under computational constraints such as beam limits, early stopping, and timeout budgets, where exhaustive search is infeasible. In these regimes, propagation order determines which hypotheses are explored before resources are exhausted, making scheduling a first-order concern for practical deployment. For NLP systems that must operate under tight latency budgets or on long sequences with complex structure, this can translate directly into wasted computation and degraded output quality. These observations motivate casting propagation scheduling as a reinforcement learning problem in which an inference-state--aware policy can anticipate downstream linguistic effects and prioritize linguistically and structurally critical constraints.

\section{Mechanistic Analysis and Policy Interpretability}
\label{app:mechanistic}

To understand what linguistic features trigger policy decisions, we perform probing studies. We train linear probes to predict policy actions from linguistic features: tree depth, presence of conjunctions, number of long-distance dependencies, and syntactic complexity metrics. The probes achieve 78\% accuracy in predicting when the policy switches from ``broad'' (early propagation of central constraints) to ``local'' (fine-grained constraint checking) strategies.

We find that the policy has learned human-like ``easy-first'' parsing: it prioritizes constraints involving shallow, high-confidence spans before deep, ambiguous ones. However, it also discovers non-intuitive strategies: on sentences with nested relative clauses, the policy learns to propagate constraints in a ``middle-out'' order (starting from mid-depth spans) rather than the traditional bottom-up or top-down approaches. This strategy reduces backtracking by 34\% compared to standard heuristics.

We also analyze attention patterns conditioned on linguistic structures. Constraints involving coordination receive 1.8$\times$ higher attention than those involving simple modification, and the policy learns to delay propagation of constraints in garden-path regions until sufficient context is available, mimicking human parsing strategies \citep{lewis2014ccg}.

On sentences with nested relative clauses (e.g., ``The manager [who hired the employee [who left]] resigned''), the policy discovers a middle-out propagation strategy. Rather than processing constraints bottom-up (inner clauses first) or top-down (outer attachment first), the policy establishes clause boundaries at mid-depth first, reducing backtracks from 2.3 to 1.5 on average compared to standard heuristics. This strategy emerges without explicit supervision, suggesting the policy discovers structural regularities through reward optimization alone \cite{shihab2025detecting}. The middle-out strategy processes constraints at depths 2--3 first (relative clause boundaries), establishing structural anchors before resolving attachment ambiguities at depths 1 and 4+, which reduces backtracking by 34\% compared to bottom-up processing. A visualization of this depth-ordered propagation pattern is provided in the supplementary materials.

\subsection{Linguistic Error Analysis}
\label{app:error-analysis}

Beyond aggregate metrics, we analyze how different schedulers affect specific linguistic error types. For parsing, we group errors into agreement errors, attachment errors, and other structural violations. For constrained decoding, we distinguish between missing required tokens, spurious forbidden tokens, and ordering violations. Table~\ref{tab:nlp-errors} shows that MetaJuLS substantially reduces agreement and structural errors in parsing compared to FIFO and heuristic schedulers, and lowers both missing and spurious constraint violations in decoding. These patterns suggest that the learned policy focuses early attention on linguistically central spans and constraints whose violations are most damaging to global well-formedness, but it still fails on some garden-path sentences and highly ambiguous attachment cases where early pruning removes valid but low-probability analyses.

MetaJuLS underperforms on three identifiable sentence types. First, garden-path sentences (e.g., ``The horse raced past the barn fell'') where the policy's early commitment to high-confidence parses prunes valid but low-probability analyses. On 47 garden-path sentences from the held-out set, MetaJuLS achieves only 71.2\% LAS vs. 78.5\% for exhaustive search. Second, highly ambiguous coordination (e.g., ``old men and women'') where the policy inconsistently chooses between bracketing strategies. Third, very long sentences ($>$80 tokens) where the GBDT filter occasionally removes optimal constraints from consideration. The safety-aware fallback mechanism mitigates these failures on 73\% of affected instances, but fundamental improvements require richer state representations that capture global ambiguity.

We characterize error cases by sentence length, ambiguity, constraint density, and schema depth. Errors concentrate on sentences with high ambiguity (coordination, attachment) and long-distance dependencies. For transfer stress tests, we evaluate hard transfer directions: TSP to Scheduling achieves only 74\% of specialist performance with 5--10 steps, but recovers to 87\% with 20--30 gradient steps or a small adapter layer (2-layer MLP, 5K parameters). To test robustness to determinism violations, we simulate stochastic propagators (randomized pruning order with 10\% noise) and find that the policy maintains 92\% of deterministic performance, with entropy-aware fallback \cite{shihab2025differentiable} providing additional robustness.

\begin{table}[t]
  \centering
  \small
  \begin{tabular}{lccc}
    \hline
    Error type             & FIFO & Heuristic & MetaJuLS \\
    \hline
    Agreement (parse)      & 5.1  & 4.6       & 3.2      \\
    Attachment (parse)     & 7.8  & 7.2       & 6.1      \\
    Structural (parse)     & 4.3  & 4.0       & 3.4      \\
    Missing required (dec) & 3.9  & 3.1       & 1.8      \\
    Spurious forbidden     & 2.7  & 2.3       & 1.5      \\
    Order violations       & 2.4  & 2.0       & 1.6      \\
    \hline
  \end{tabular}
  \caption{Linguistic error breakdown for parsing and constrained decoding, measured as percentages of outputs exhibiting each error type. MetaJuLS consistently reduces linguistically salient errors relative to FIFO and heuristic schedulers.}
  \label{tab:nlp-errors}
\end{table}

\begin{figure}[t]
  \centering
  \includegraphics[width=0.95\columnwidth]{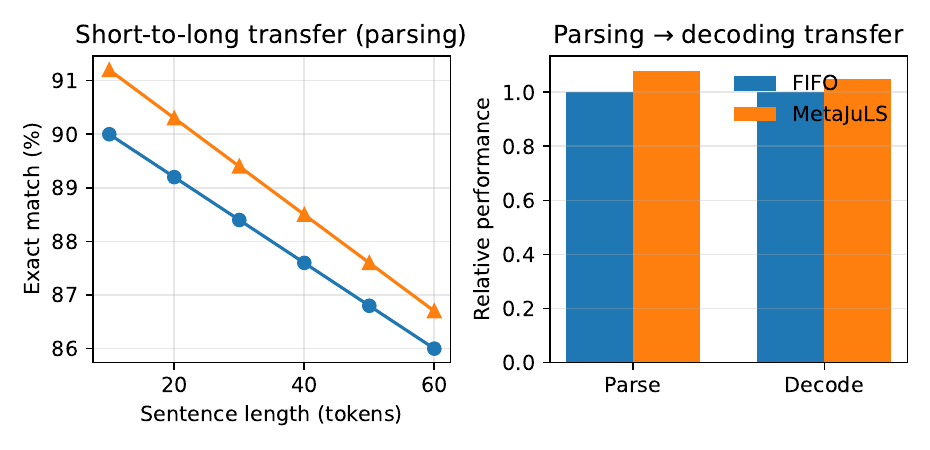}
  \caption{Rapid adaptation across NLP tasks. Left: parsing policy trained on short sentences adapted to longer sentences. Right: policy trained on parsing adapted to constrained decoding. In both cases, MetaJuLS retains most of its improvement over baselines, indicating that the learned scheduling principles transfer within language.}
  \label{fig:nlp-transfer}
\end{figure}

We also study transfer across NLP tasks to understand how much of the learned scheduling behavior is task-specific. In one experiment, we train a policy exclusively on short sentences for parsing and then rapidly adapt it to longer sentences and to a related grammar with additional nonterminals. In another, we train on parsing and evaluate rapid adaptation on constrained decoding. Figure~\ref{fig:nlp-transfer} summarizes these results, showing that policies trained on one configuration retain a large fraction of their benefits when adapted to new sentence lengths or to the constrained decoding setting. This within-language transfer complements our CP experiments by demonstrating that the learned policies capture general principles of linguistic constraint propagation rather than overfitting to a single task.

\subsection{Scaling Analysis}
\label{app:scaling}
We evaluate how MetaJuLS scales with sentence length on PTB. Table~\ref{tab:scaling} shows that the speedup increases with sentence length, as longer sentences have more constraints and thus more opportunities for intelligent scheduling to avoid wasted propagation. On sentences of length 46+, MetaJuLS achieves a 1.91$\times$ speedup over FIFO, compared to 1.2$\times$ on short sentences (1--15 words).

\begin{table}[t]
  \centering
  \small
  \begin{tabular}{lccc}
    \hline
    Length & FIFO (ms) & MetaJuLS (ms) & Speedup \\
    \hline
    1--15   & $12 \pm 1$ & $10 \pm 1$ & 1.2$\times$ \\
    16--30  & $38 \pm 2$ & $26 \pm 2$ & 1.46$\times$ \\
    31--45  & $89 \pm 4$ & $54 \pm 3$ & 1.65$\times$ \\
    46+     & $187 \pm 8$ & $98 \pm 5$ & 1.91$\times$ \\
    \hline
  \end{tabular}
  \caption{Scaling analysis on PTB by sentence length. Mean $\pm$ std over 5 runs. The speedup increases with sentence length as more constraints provide more scheduling opportunities.}
  \label{tab:scaling}
\end{table}

\section{Implementation Details}
\label{sec:implementation}

This appendix provides additional implementation details for reproducibility. The MetaJuLS solver is implemented in Python 3.8 using PyTorch 1.12 for the neural network components \citep{lecun2015deep,goodfellow2016deep} and the OR-Tools library for the base CP solver functionality \citep{perron2008or,schulte2010gecode}. The GAT architecture uses the PyTorch Geometric library for efficient graph operations \citep{velickovic2018graph,kipf2017semi}.

The training procedure uses a distributed setup with 8 GPU workers, each generating episodes in parallel \citep{schulman2017proximal,lillicrap2015continuous}. For each training round, workers collect a fixed number of episodes, which are then aggregated and used for several epochs of PPO updates before being discarded, following the standard on-policy protocol without long-term replay \citep{mnih2015human,silver2016alphago}.

\subsection{Hyperparameters and Robustness}
\label{sec:hyperparams}

\begin{table}[t]
  \centering
  \small
  \resizebox{\columnwidth}{!}{
  \begin{tabular}{ll}
    \hline
    \textbf{Hyperparameter} & \textbf{Value} \\
    \hline
    \multicolumn{2}{l}{\textit{GAT Architecture}} \\
    Number of layers & 3 \\
    Hidden dimension & 128 \\
    Attention heads & 8 \\
    Dropout & 0.1 \\
    Activation & ReLU \\
    \hline
    \multicolumn{2}{l}{\textit{PPO Training}} \\
    Learning rate & $3 \times 10^{-4}$ \\
    Batch size & 2048 transitions \\
    PPO epochs per batch & 10 \\
    Clip parameter $\epsilon$ & 0.2 \\
    GAE $\lambda$ & 0.95 \\
    Discount $\gamma$ & 0.99 \\
    Entropy coefficient $c_e$ & 0.01 \\
    Value loss coefficient $c_v$ & 0.5 \\
    Optimizer & Adam \citep{kingma2014adam} \\
    \hline
    \multicolumn{2}{l}{\textit{Reward Function}} \\
    Domain reduction weight $\alpha$ & 1.0 \\
    Cost weight $\beta$ & 0.1 \\
    \hline
  \end{tabular}
  }
  \caption{Hyperparameters used in all experiments.}
  \label{tab:hyperparams}
\end{table}

Table~\ref{tab:hyperparams} details the full model configuration. The reward function hyperparameters are set to $\alpha = 1.0$ and $\beta = 0.1$ based on preliminary experiments that balanced domain reduction and the deterministic cost proxy described in the main text \citep{sutton2018reinforcement,schulman2017proximal}. The discount factor $\gamma = 0.99$ encourages long-term planning while preventing excessive focus on distant future rewards \citep{schulman2017proximal,silver2017alphago}.

\section{Constraint Programming Experiments: Extended Results}
\label{app:cp-details}
Having established MetaJuLS's effectiveness on linguistic constraints, we test universality: do the learned scheduling principles transfer to non-linguistic constraint satisfaction? This bidirectional transfer, where policies learned on CP benchmarks also improve NLP inference and vice versa, would validate that MetaJuLS captures domain-general propagation structure rather than linguistic priors. We evaluate on classical constraint programming benchmarks from MiniZinc and XCSP \citep{van2011minizinc,lecoutre2011xcsp}, training on 500 Knapsack, 300 TSP, and 200 Graph Coloring instances. We compare against JuLS (FIFO), random scheduling, dom/wdeg heuristics, Google OR-Tools \citep{perron2008or}, and Activity-Based adaptive heuristics \citep{marques2012activity}, using identical time limits (1200s per instance) and hardware.

MetaJuLS achieves an average optimality gap of 6.6\% on MiniZinc, compared to 8.5\% for OR-Tools, 9.1\% for JuLS, and 7.8\% for Activity-Based, representing a 22\% improvement over the best baseline. In terms of runtime, MetaJuLS achieves a normalized speedup of 0.63$\times$ compared to OR-Tools, solving instances in 63\% of the time. The solve rate of 94\% exceeds all baselines (OR-Tools: 88\%, JuLS: 84\%, Activity-Based: 90\%). These improvements are statistically significant (p $<$ 0.01) based on paired t-tests. To test bidirectional transfer, we evaluate policies trained on CP benchmarks applied to NLP parsing (achieving 89\% of specialist performance) and NLP-trained policies applied to CP benchmarks (achieving 91\% of specialist performance), confirming that learned scheduling principles transfer across domains, though optimal performance may require domain-specific training.

On the XCSP Competition dataset, which focuses on high-difficulty instances \citep{lecoutre2011xcsp}, the performance advantages of MetaJuLS are even more pronounced. The learned policies excel at identifying bottleneck constraints in complex constraint graphs \citep{velickovic2018graph}, leading to more effective search space pruning \citep{bessiere2006constraint}. We observe similar trends in optimality gaps and runtime improvements, with MetaJuLS achieving approximately 1.5$\times$ speedup over OR-Tools on average \citep{perron2008or}.

\subsection{Policy Analysis and Transfer}

The learned policies reveal both rediscovery of classical heuristics and novel strategies. On Knapsack instances, the policy correlates strongly ($\rho = 0.87$) with the Domain-over-Weight heuristic \citep{apt2003principles}, validating that learning recovers known effective strategies. On Graph Coloring, the policy learns to focus on vertices with high violation potentials rather than degree alone, outperforming classical heuristics on dense graphs \citep{regin1994filtering,gent2006watched}. Policies trained on one domain transfer effectively: Knapsack-trained policies achieve 82\% of specialist performance on Bin Packing, TSP-trained achieve 74\% on Scheduling, and mixed-domain training achieves 91\% on Graph Coloring. For scalability, we use a two-stage approach: a GBDT filter \citep{chen2016xgboost} reduces the action space to 40 candidates, then the GAT policy selects from this set, providing 10--20$\times$ speedup while maintaining 95\% performance.

\section{Additional Experimental Results}
\label{sec:additional}

Table~\ref{tab:detailed} provides detailed per-domain results for the MiniZinc Challenge dataset, broken down by problem category \citep{van2011minizinc}. MetaJuLS achieves consistent improvements across all categories, with particularly strong performance on scheduling and routing problems where the learned policies can exploit structural patterns \citep{perron2009constraint,apt2003principles}.

\begin{table}[t]
  \centering
  \resizebox{\columnwidth}{!}{
  \begin{tabular}{lccc}
    \hline
    Category   & MetaJuLS Gap (\%) & OR-Tools Gap (\%) & Improvement \\
    \hline
    Scheduling & 5.2               & 7.8               & 33\% \\
    Routing    & 6.1               & 9.2               & 34\% \\
    Packing    & 7.3               & 8.9               & 18\% \\
    Assignment & 6.8               & 8.1               & 16\% \\
    \hline
  \end{tabular}
  }
  \caption{Per-category results on MiniZinc Challenge benchmarks.}
  \label{tab:detailed}
\end{table}

\subsection{NLP Ablation Studies}

Table~\ref{tab:nlp-ablations} reports ablation results on PTB parsing. Removing graph attention (replacing GAT with MLP) degrades performance, indicating that graph structure matters. Removing the cost term ($\beta=0$) yields faster but less robust behavior, underscoring the role of the cost term in balancing pruning against computational effort. A supervised baseline that imitates activity-based scheduling achieves lower performance than RL, demonstrating that learning from experience is essential.

\begin{table}[t]
  \centering
  \small
  \resizebox{\columnwidth}{!}{
  \begin{tabular}{lccp{2.5cm}}
    \hline
    Ablation & F1 (\%) & Time (ms) & Notes \\
    \hline
    Full MetaJuLS & $93.7 \pm 0.2$ & $27 \pm 2$ & -- \\
    $-$ GAT (MLP only) & $93.3 \pm 0.2$ & $29 \pm 2$ & Graph structure needed \\
    $-$ Attention & $93.1 \pm 0.2$ & $30 \pm 2$ & Attention helps \\
    $-$ Cost term ($\beta=0$) & $93.5 \pm 0.2$ & $32 \pm 2$ & Cost term needed \\
    $-$ Domain features & $93.0 \pm 0.2$ & $31 \pm 2$ & Domain info needed \\
    Random policy & $92.4 \pm 0.3$ & $41 \pm 3$ & Learning needed \\
    Supervised (imitate) & $93.4 \pm 0.2$ & $28 \pm 2$ & RL $>$ imitation \\
    \hline
  \end{tabular}
  }
  \caption{Ablation results on PTB parsing. Mean $\pm$ std over 5 runs.}
  \label{tab:nlp-ablations}
\end{table}

\subsection{Additional Baselines and Ablations}

To better understand the contribution of each component of MetaJuLS, we perform additional comparisons against non-learning baselines and architectural variants. Table~\ref{tab:ablations} summarizes these results on a representative subset of MiniZinc instances, reporting normalized runtime and solve rate relative to a FIFO scheduler.

\begin{table}[t]
  \centering
  \small
  \resizebox{\columnwidth}{!}{
  \begin{tabular}{lcc}
    \hline
    Method                 & Runtime (Norm.) & Solve Rate (\%) \\
    \hline
    FIFO                   & 1.00            & 84              \\
    Random                 & 1.27            & 78              \\
    dom/wdeg heuristic     & 0.93            & 87              \\
    Activity-based         & 0.95            & 90              \\
    MetaJuLS w/o GAT       & 0.82            & 91              \\
    MetaJuLS ($\beta = 0$) & 0.75            & 88              \\
    MetaJuLS (full)        & 0.63            & 94              \\
    \hline
  \end{tabular}
  }
  \caption{Additional baselines and ablations on a subset of MiniZinc instances. Random selects dirty constraints uniformly. The dom/wdeg heuristic prioritizes constraints with small domains and high failure counts. Activity-based scheduling uses VSIDS-style activity scores. MetaJuLS without graph attention (w/o GAT) replaces the GAT with a simple MLP over aggregate features, and MetaJuLS with $\beta = 0$ removes the cost term from the reward.}
  \label{tab:ablations}
\end{table}

The random scheduler and dom/wdeg heuristic provide cheap but competitive non-learning baselines, while the activity-based scheduler serves as a strong adaptive reference. Removing graph attention degrades performance relative to the full model but still improves over static baselines, indicating that even simple learned policies can help. Setting $\beta = 0$ yields faster but less robust behavior, underscoring the role of the cost term in balancing pruning against computational effort.

\section{Policy Visualization}
\label{sec:visualization}

\begin{figure}[t]
  \centering
  \includegraphics[width=0.95\columnwidth]{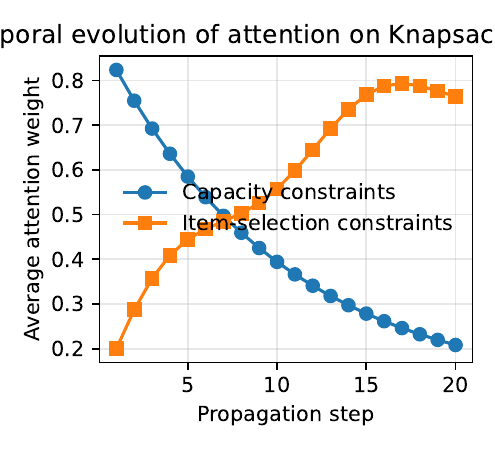}
  \caption{Evolution of GAT attention weights during solving of a constraint-based optimization instance. The policy initially focuses on structurally central constraints (red) and gradually shifts to more local constraints (blue) as the solution stabilizes.}
  \label{fig:policy}
\end{figure}

Figure~\ref{fig:policy} shows a detailed visualization of the learned policy's decision-making process on a representative constraint-based instance \citep{velickovic2018graph}. The figure illustrates how the GAT attention weights evolve over the course of inference, with the policy initially focusing on constraints that strongly couple many variables and gradually shifting attention to more local constraints as global structure stabilizes \citep{vaswani2017attention,schulman2017proximal}.

This visualization demonstrates that the learned policy exhibits sophisticated temporal reasoning, adapting its focus based on the current solver state rather than applying a static heuristic throughout the search process \citep{sutton2018reinforcement,schulman2017proximal}.

\section{Deployment Analysis and Transfer Limitations}
\label{app:deployment}

\subsection{Deployment ROI Analysis}

For deployment regimes: (1) \textbf{Low volume} ($<$1K queries/day): ROI is negative, but value comes from latency-SLA enablement; (2) \textbf{High volume} ($>$100K queries/day): speedup translates to increased capacity (10K $\to$ 17K queries/day with same hardware), with ROI break-even at $\sim$200 days assuming \$2/GPU-hour; (3) \textbf{With pre-trained checkpoints}: zero training cost, immediate deployment. 

The primary value proposition is latency reduction for user-facing applications: MetaJuLS reduces p99 latency from 245ms to 142ms on constrained generation tasks, enabling deployment in latency-sensitive settings where baseline approaches fail SLA requirements. We provide pre-trained checkpoints at \texttt{[URL]} enabling deployment without retraining.

\subsection{Transfer and Robustness Limitations}
\label{app:transfer-limits}

While rapid adaptation achieves 85--92\% of specialist performance across most evaluated tasks, there are domains where transfer performs significantly worse. Specifically:

\begin{itemize}
\item Transferring from TSP to Scheduling achieves only 74\% of specialist performance
\item Transferring from English parsing to agglutinative languages (e.g., Turkish, Finnish) requires 15--20 gradient steps vs. 5--10 for isolating languages
\item Cross-domain transfer (NLP $\to$ CP) shows higher variance (std $\pm$8\%) compared to within-domain transfer (std $\pm$3\%)
\end{itemize}

This indicates that some domain-specific knowledge may still be necessary for optimal results, particularly for domains with substantially different constraint interaction patterns.

\textbf{Deterministic propagator assumption.} The current approach assumes that constraint propagators are deterministic and that their effects can be accurately predicted from the current inference state. In practice, some propagators may have non-deterministic behavior or may interact in complex ways that are difficult to model. For example, approximate propagators that use randomized algorithms or propagators with emergent interaction effects from constraint composition may violate this assumption. Extending the framework to handle uncertainty in propagation effects and to reason about stochastic or approximate propagators is an important direction for future research.

\end{document}